%% file: preprint.tex
\documentclass{article}

\usepackage{arxiv}

\usepackage[utf8]{inputenc}
\usepackage[T1]{fontenc}
\usepackage{hyperref}
\usepackage{url}
\usepackage{booktabs}
\usepackage{amsfonts}
\usepackage{nicefrac}
\usepackage{microtype}
\usepackage{lipsum}
\usepackage{graphicx}
\usepackage[round,authoryear]{natbib}
\usepackage{doi}
\usepackage{enumitem}
\usepackage{pdflscape}
\usepackage{amsmath}

\usepackage{xltabular}
\usepackage{multirow} 
\usepackage[table]{xcolor}
\usepackage{caption}
\title{STeMP: Spatio-Temporal Modelling Protocol}

\author{ \href{https://orcid.org/0000-0003-0991-8646}{\includegraphics[scale=0.06]{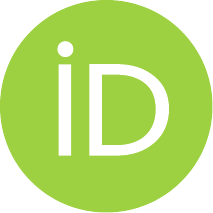}\hspace{1mm}Jan Linnenbrink} \\
	Institute of Landscape Ecology\\
	University of Münster\\
	48149 Münster, Germany \\
	\texttt{jan.linnenbrink@uni-muenster.de}\\
	\And
	\href{http://orcid.org/0000-0002-1057-3721}{\includegraphics[scale=0.06]{orcid.pdf}\hspace{1mm}Jakub Nowosad} \\
	Institute of Landscape Ecology\\
	University of Münster\\
	48149 Münster, Germany \\
    Institute of Geoecology and Geoinformation\\
	Adam Mickiewicz University\\
	61-680 Poznań, Poland \\
    \And
    \href{https://orcid.org/0000-0002-3010-018X}{\includegraphics[scale=0.06]{orcid.pdf}\hspace{1mm}
    Marvin Ludwig} \\
	Institute of Landscape Ecology\\
	University of Münster\\
	48149 Münster, Germany \\
	\And
    {Anna Frederike Jablotschkin} \\
	Institute of Landscape Ecology\\
	University of Münster\\
	48149 Münster, Germany \\
	\And
    {Fabian Schumacher} \\
	Institute of Landscape Ecology \\
	University of Münster \\
	48149 Münster, Germany \\
	\And
    \href{https://orcid.org/0000-0001-7381-3828}{\includegraphics[scale=0.06]{orcid.pdf}\hspace{1mm}Teja Kattenborn} \\
	Chair of Sensor-based Geoinformatics (geosense) \\
	University of Freiburg\\
	79106 Freiburg, Germany \\
	\And
    \href{https://orcid.org/0000-0003-0556-0210}{\includegraphics[scale=0.06]{orcid.pdf}\hspace{1mm}Hanna Meyer} \\
	Institute of Landscape Ecology\\
	University of Münster\\
	48149 Münster, Germany \\
	\texttt{hanna.meyer@uni-muenster.de} \\
}

\renewcommand{\shorttitle}{STeMP}

\hypersetup{
pdftitle={STeMP},
pdfauthor={Jan Linnenbrink, Jakub Nowosad, Marvin Ludwig, Anna Frederike Jablotschkin, Fabian Schumacher, Teja Kattenborn, Hanna Meyer},
pdfkeywords={machine-learning, model protocol, model report, reproducibility, spatial prediction, transparency},
}

\begin{document}
\maketitle

\begin{abstract}
	1. Spatio-temporal predictive machine-learning modelling is an important tool in environmental research and the geosciences. However, machine-learning models are highly sensitive to both the characteristics of the training data, such as its distribution, and methodological choices, including the cross-validation strategy and model tuning. Each decision has impact and implications on the model itself as well as the estimation of the model quality and applicability for certain purposes. Taking into account the large role of machine-learning based maps of the environment in science and their transfer into practice, transparent reporting of spatio-temporal models, ideally using standardized model protocols, is essential to enable trust, transparency and comparability. However, such protocols are currently lacking for spatio-temporal modelling.

    2. We propose STeMP (Spatio-Temporal Modelling Protocol) to fill this gap by serving two purposes: standardized reporting to understand the model functioning as well as providing guidance during the modelling process by pointing at critical decisions and parameters. 
    The protocol is structured in three sections: Overview, Model and Prediction. The Overview section contains metadata, while the Model and Prediction sections go into detail, describing predictors, evaluation and software, and further relevant elements of the modelling workflow.

    3. The protocol definition is hosted on GitHub and accompanied by an R-package (\url{https://github.com/LOEK-RS/STeMP}). The R-package contains a web application that can be used to fill the protocol either manually or in a semi-automated way from provided modelling objects. Warnings are returned from the protocol when common pitfalls are encountered, which may help authors as a guide through the modelling process but also support reviewers in the assessment of predictive modelling studies. Via GitHub, incorporation of contributions and feedback from the community is encouraged.
    
\end{abstract}

\keywords{machine-learning \and model protocol \and model report \and reproducibility \and spatial prediction \and transparency}

\section{Introduction}

Machine-learning methods have become increasingly popular in geoscientific spatio-temporal modelling. They allow to learn complex, non-linear and often interacting relationships without the need to make assumptions about the distribution of the data, as in statistical modelling. 
This leads to applications of machine-learning in diverse geoscientific fields such as climatology \citep{Yang2024}, ecology \citep{Pichler2023}, remote sensing and soil sciences \citep{Wadoux2025}.
However, machine-learning models are often perceived as "black boxes", meaning that it is hard to understand their learned relationships. Furthermore, several modelling decisions must be made that may considerably change the model, the resulting performance estimates, and its interpretation. This applies to machine-learning models in general, but spatio-temporal machine-learning models in particular. Such issues may, for example, concern the autocorrelation inherent in spatio-temporal data, which complicates data splitting for model evaluation and explanation. Failure to account for these effects can lead to substantial pitfalls like over-optimistic performance estimates, which may result in false confidence in the models and may ultimately even harm the trust in spatial predictive modelling in general \citep{Nowosad2026}.

Transparently reporting important modelling decisions, like data splitting and model selection, as well as input data characteristics is thus key so that practitioners using the model and reviewers assessing it can understand its limitations, and to evaluate if they should trust the predictions made for a given purpose (e.g., for specific regions or times or where a particular quality is required). It also helps authors of a spatio-temporal machine-learning model to explicitly evaluate their modelling decisions, and to adhere to best practices. Furthermore, while a model has usually been tied to a specific prediction in environmental modelling, there is a recent shift towards the publication of models that might be applied for a wider range of tasks \citep{Graser2026}. Transparent reporting is essential in this case to enable the user of the model to assess if a model is suitable for a particular task (e.g. specific region or time).
However, such transparency is often lacking in spatio-temporal machine-learning studies in geoscientific applications. For example, information on the resolution of the predictors, the distribution of the training points, and the chosen evaluation strategy is often not transparently communicated, or they are scattered through the manuscript and the appendix \citep[see ][]{Zurell2020}. Such information, however, is essential to interpret the models' applicability for a given task, as well as the reliability of the communicated performance statistics.

Standardized model protocols are helpful in achieving transparency and reproducibility in machine-learning studies, and thus facilitate review of models, interpreting their outcomes, and enabling their reproduction. Some model protocols exist for machine-learning models in general \citep[e.g.,][]{Mitchell2019}, and there also exist a large number of protocols in fields subject to substantial legal and institutional oversight, such as medicine \citep[e.g.,][]{Collins2015, Mongan2020, Luo2016}. In environmental modelling, there are existing definitions of model protocols for more specific applications. The ODMAP protocol \citep{Zurell2020} was developed to describe species distribution models, while the ODE protocol \citep{Seuru2026} is directed towards socio-ecological studies. Both build on the ODD protocol \citep{Grimm2006}, which was designed to report individual-based models in ecology. In addition to model protocols, software checklists focusing on best practices regarding the functional design, objects and naming conventions exist for (spatial) machine-learning and statistical models \citep[e.g. the rOpenSci Statistical Software Peer Review,][]{Ram2019}. However, a standardized model protocol focusing on the field of spatio-temporal machine-learning modelling, with its unique characteristics such as spatial autocorrelation, is missing \citep{Graser2026}. This hinders transparency and misses the chance to detect and mitigate pitfalls specific to spatio-temporal machine-learning, such as inadequate data splitting or predicting outside the area where the model is applicable \citep{Nowosad2026}. Hence, such a standardized protocol tailored towards spatio-temporal machine-learning models in environmental sciences is needed to fill the gap between general machine-learning protocols and protocols specialized on very specific fields of environmental mapping. 

Here we propose STeMP, a \textbf{S}patio-\textbf{Te}mporal \textbf{M}odelling \textbf{P}rotocol, to fill this gap. It focuses on both, the model properties -- which are important for researchers that aim to apply the model to their own task -- and the documentation of the prediction, which might be more important for practitioners or scientists working with the model output (e.g. the map). It covers crucial properties of the data, which have important implications for the modelling process. STeMP is accompanied by an R-package containing a web application and R-functions to analyze protocol outputs. While the protocol and the software benefit several stakeholders involved in the modelling process, it incorporates specific features tailored to three primary groups:

\begin{itemize}
\item \emph{Model developers}: The web application provides the functionality to spot potential issues specified by the protocol and communicate them as warnings, allowing developers to proactively address and mitigate potential errors well before peer-review or deployment.
\item \emph{Model evaluators}: R-functions accompanying STeMP provide automated summaries of identified issues that enable reviewers to efficiently evaluate the most critical aspects of the model.
\item \emph{Model users}: Detailed documentation allows end-users to more easily assess the model's suitability for a specific context and compare it with alternative models.
\end{itemize}

Community involvement is key to the long-term success of such a protocol. Thus, we maintain the definition of the protocol and the accompanying software on GitHub (\url{https://github.com/LOEK-RS/STeMP}), open for suggestions or discussion from the community.

\section{Conceptual Design} \label{sec:elements}

\begin{figure}
    \centering
    \includegraphics[width=0.8\linewidth]{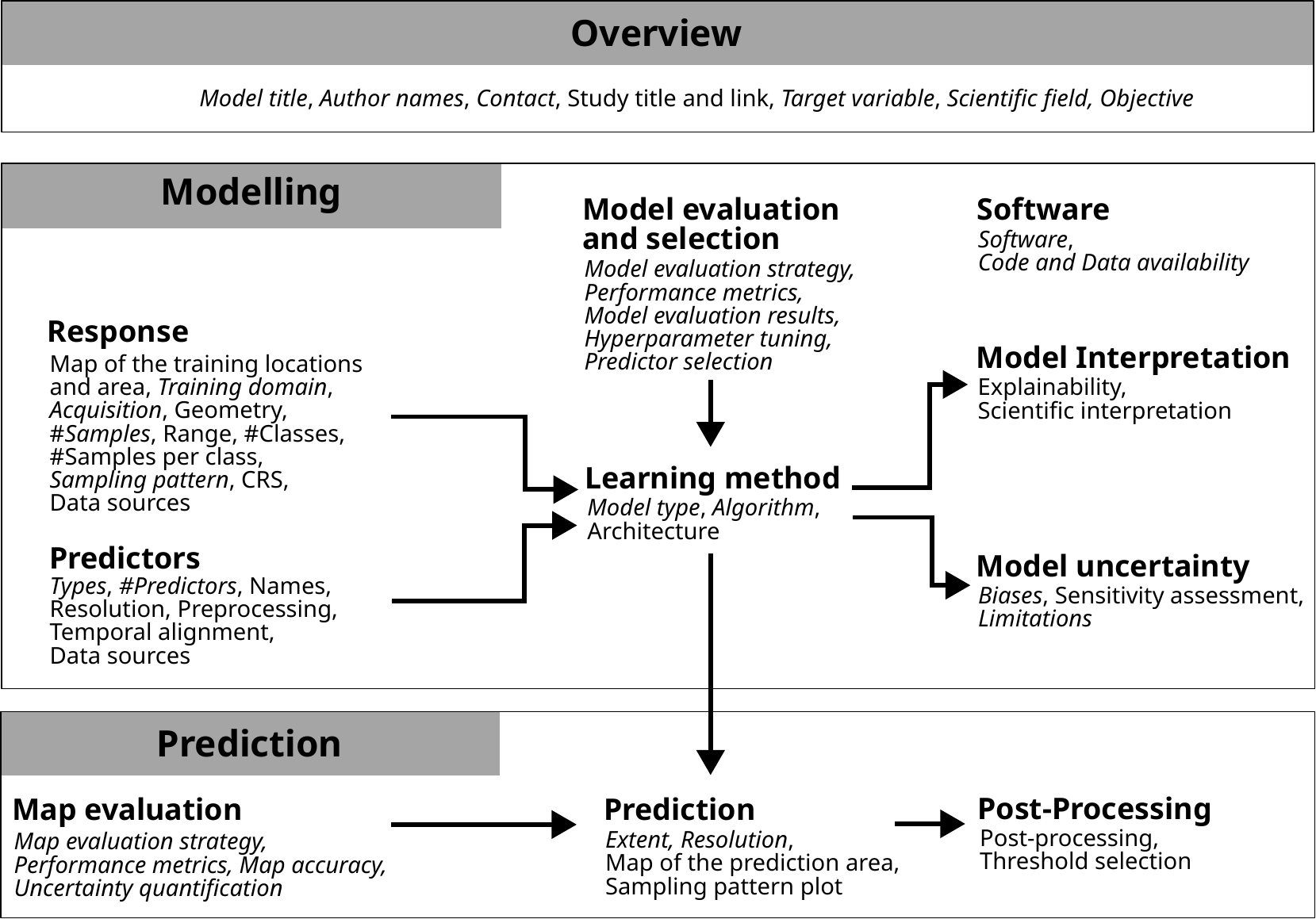}
    \caption{The structure of STeMP. The three boxes correspond to the three sections of the protocol. The protocol is aligned to typical machine-learning workflows, which start with the response and predictor data acquisition and end with the post-processing of the predictions. Bold text shows the sub-sections of the protocol, while italic text refers to mandatory fields. See Table \ref{tab:model_elements} for a complete list of all fields and corresponding short descriptions.}
    \label{fig:overview}
\end{figure}

Aligned with standard spatio-temporal machine-learning workflows, we structured STeMP into three main sections: \emph{Overview}, \emph{Model} and the optional \emph{Prediction} section (Figure \ref{fig:overview} \& Table \ref{tab:model_elements}). The \emph{Overview} section provides general information about the model (such as the author names, contact information, and if there is a publication associated with the model). The \emph{Model} and \emph{Prediction} sections document the decisions taken in the modelling process and the corresponding predictions.
The \emph{Model} section is organized in seven sub-sections that correspond to the principal stages of the modelling workflow. It covers information about the learning method -- specifying the algorithm employed and whether it was used for classification or regression -- together with the characteristics of the response data like the sampling design, as well as spatial and temporal resolution of the predictor variables. The section also addresses the model evaluation and selection, the interpretation of the model based on explainable AI techniques and details on potential biases or limitations of the model. Technical aspects such as the software implementation of the model can also be specified.
The \emph{Prediction} section is optional since models are not necessarily tied to a specific prediction or application, but may be developed to be used in different contexts and by different end-users. The prediction section covers three sub-sections: the prediction domain, the final map evaluation and uncertainty assessment, and lastly the post-processing of the resulting map. Thereby, it covers information on the method used for uncertainty quantification (e.g., quantile random forest \citep{Meinshausen2006} to quantify the prediction interval, or AOA \citep{Meyer2021} and MESS \citep{Elith2010} to delineate areas not covered by the training data), the spatio-temporal dimensions of the prediction domain, as well as post-processing measures such as threshold selection when a continuous prediction needs to be binarized.

\section{Software accompanying STeMP}

\subsection{Functionality}

STeMP is accompanied by an R-package that contains a web application and R-functions to streamline protocol completion, increase the utility of the protocol during model development, and improve its usefulness for model reviewers.
The web application is designed to meet the needs of the majority of users, while the R functions offer additional functions to read and analyze completed protocols.

The web application allows many fields to be automatically filled by retrieving information from uploaded models and spatial datasets. At the moment, model objects stored as RDS objects and trained using the R-packages \texttt{caret} \citep{caret}, \texttt{tidymodels} \citep{tidymodels} and \texttt{mlr3} \citep{mlr3} are supported. From these objects, information about sample size, response characteristics, model parameters and others are extracted and the corresponding fields of the protocol are automatically filled. Moreover, spatial datasets can be uploaded, including the training locations and the prediction area. From these datasets, information such as the coordinate reference system can be extracted, and maps of the training locations and prediction area are automatically generated. When both training locations and prediction area are uploaded, the sampling pattern is automatically inferred based on spatial nearest-neighbour distances (see \citealt{Meyer2026, Linnenbrink2024, Baddeley2015} for more details).

In addition to automation, the web application raises warnings when common spatial machine-learning pitfalls are detected \citep{Nowosad2026}. Currently, the following aspects are checked:

\begin{itemize}
    \item Evaluation strategies not suited to the prediction task (e.g., random cross-validation with spatially clustered training points).
    \item Data leakage between training and evaluation of models.
    \item The absence of uncertainty quantification when the models are likely applied in extrapolation situations (i.e., when the training points are clustered).
    \item The use of spatial proxies where not appropriate, or without rigorous predictor selection.
\end{itemize}

While detected issues should not automatically be interpreted as errors -- as they may result from conscious and well-motivated modelling decisions -- they still warrant careful consideration or justification.
Providing warnings about these potential issues makes the protocol useful during model development: when the protocol is completed already at this stage, potential problems can be identified and addressed early in the modelling process.

Finally, the utility of the web application is extended by the R functions accompanying it. Most importantly, the function \texttt{protocol\_analyze} generates a report from a downloaded protocol. This report summarizes all persisting warnings that were raised when filling the protocol, hence providing a concise summary of potential issues encountered during the model building process. This may be particularly useful for reviewers of a model, allowing them to quickly identify and evaluate aspects requiring closer scrutiny.

\subsection{Implementation}

STeMP is implemented as a modular Shiny web application \citep{Chang2026, Wickham2021}, which makes the reporting workflow straightforward. We built upon the ODMAP web application \citep{Zurell2020}, which we restructured and extended to accommodate the requirements of the STeMP protocol. To ensure maintainability and scalability, we developed the web application as an R-package that follows the \texttt{golem} framework \citep{Fay2026, Fay2021}. This R-package additionally contains functions to read a downloaded protocol as a \texttt{data.frame} in R (\texttt{protocol\_read}) and to summarize the warnings that were raised by the web application (\texttt{protocol\_analyze}). The software is shared together with the protocol definition on GitHub (\url{https://github.com/LOEK-RS/STeMP}) under the GNU General Public License, and the version presented in this manuscript, v2026.07.00, is archived on Zenodo (\url{https://doi.org/10.5281/zenodo.21494506}). In the following paragraphs, we describe the technical design of the web application.

The code of the STeMP web application is organized into several modules, reflecting the structure of the protocol described in section \ref{sec:elements}. This modular approach not only aligns well with the conceptual design of the protocol, but also facilitates maintenance of the web application and enables re-use of code blocks. 
At the highest level, the \emph{run\_app} function calls the two top-level modules \emph{app\_ui} and \emph{app\_server}. These top-level modules then invoke different modules to specify the user interface and the server logic. 

The modules \emph{about} and \emph{howto} define the landing pages of the web application and do not contribute server logic. The remaining modules contain server logic and, at the same time, define the user interface. The modules \emph{upload} and \emph{spatialdata\_upload} enable uploading model objects and spatial data. The modules \emph{model\_metadata} and \emph{spatialdata\_metadata} automatically extract information from the uploaded objects to update the respective fields in the protocol. The protocol itself is rendered based on the module \emph{create\_protocol}, which is divided into three submodules covering the three sections of the protocol: Overview, Model and Prediction. Downloading the protocol as a CSV or PDF file, or downloading the maps and plot of nearest-neighbour distances created by the web application is enabled through the \emph{sidebar} module. This module also allows users to upload a previously started protocol as CSV or ZIP file by calling \emph{mod\_csv\_zip\_upload}, enabling them to continue working on an existing protocol. Warning messages based on the input of the user may be raised, as defined in the \emph{warnings} module. Lastly, the \emph{viewer} module creates a preview of the protocol in HTML format.

We incorporated tests at three levels to cover all aspects of the web application. 
We utilized \texttt{testthat} \citep{Wickham2011} for unit tests of important functions like the classification of the sampling pattern. Then, we used the \texttt{testServer} function from \texttt{Shiny} to test the reactivity inside the server functions. Lastly, we employed snapshot-based testing provided by \texttt{shinytest} \citep{Chang2024} to test the general behavior of the web application, including the server functions as well as the user interface.

\section{Example} 
Here we present the application of STeMP in a typical spatial predictive modelling task. In the following description, we only present the information usually included in research manuscripts: the final prediction map and the evaluation statistic. The code and data to reproduce the example, as well as a Figure showing a more complete description of the example are available as an appendix (\url{https://doi.org/10.5281/zenodo.21493718}).

\begin{figure}
    \centering
    \includegraphics[width=0.4\linewidth]{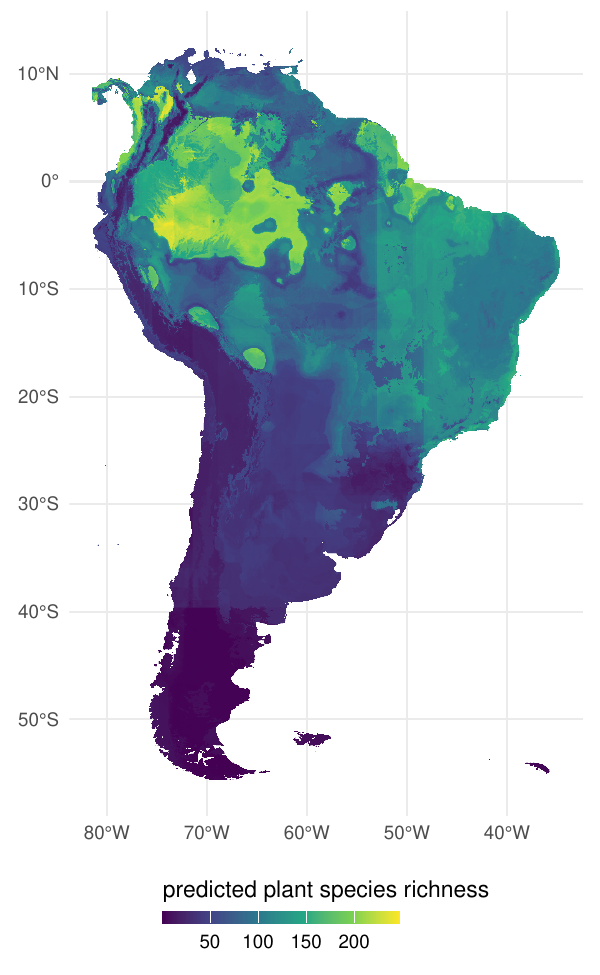}
    \caption{The prediction resulting from the example modelling task.}
    \label{fig:example_prediction}
\end{figure}

We use the prediction of plant species richness in South America as an example \citep[used to illustrate concepts in ][] {Meyer2026}. What is typically written in a scientific paper to document the modelling process and the results could be the following:\\ \textit{We trained a random forest model based on plot-based measurements of plant species richness obtained from the sPlotOpen database \citep{Sabatini2021} combined with spatially continuous predictors describing elevation \citep{Jarvis2008}, spatial location and the climate (WorldClim, \citealt{Fick2017}). The model performance was high, indicated by an $R^2$ value of 0.7, suggesting that the resulting map is suitable to be used by planning authorities, or as an input for further modelling studies (Figure \ref{fig:example_prediction} shows the predicted plant species richness for South America).}

However, there are several open questions that may come to mind for potential users of the map (or the model), as well as for researchers who want to understand the model or may want to compare different maps of plant species richness in South America generated from different models.\\
A researcher who wants to use the map as input for downstream modelling might ask:

\begin{enumerate}[series=questions]
    \item How are the training points distributed in the prediction area?
    \item Is the communicated performance statistic trustworthy, and for which area does it apply?
    \item What causes the spatial patterns in the predicted map of plant species richness?
\end{enumerate}

If the researcher would be interested in the plant diversity of North America, they might ask:
\begin{enumerate}[resume=questions]
    \item Can I use the model to make predictions for North America?
\end{enumerate}

A nature conservation organization working on a more regional scale might additionally be interested in the question:
\begin{enumerate}[resume=questions]
    \item Can I use the map for local planning?
\end{enumerate}

\begin{figure}
    \centering
    \includegraphics[width=0.8\linewidth]{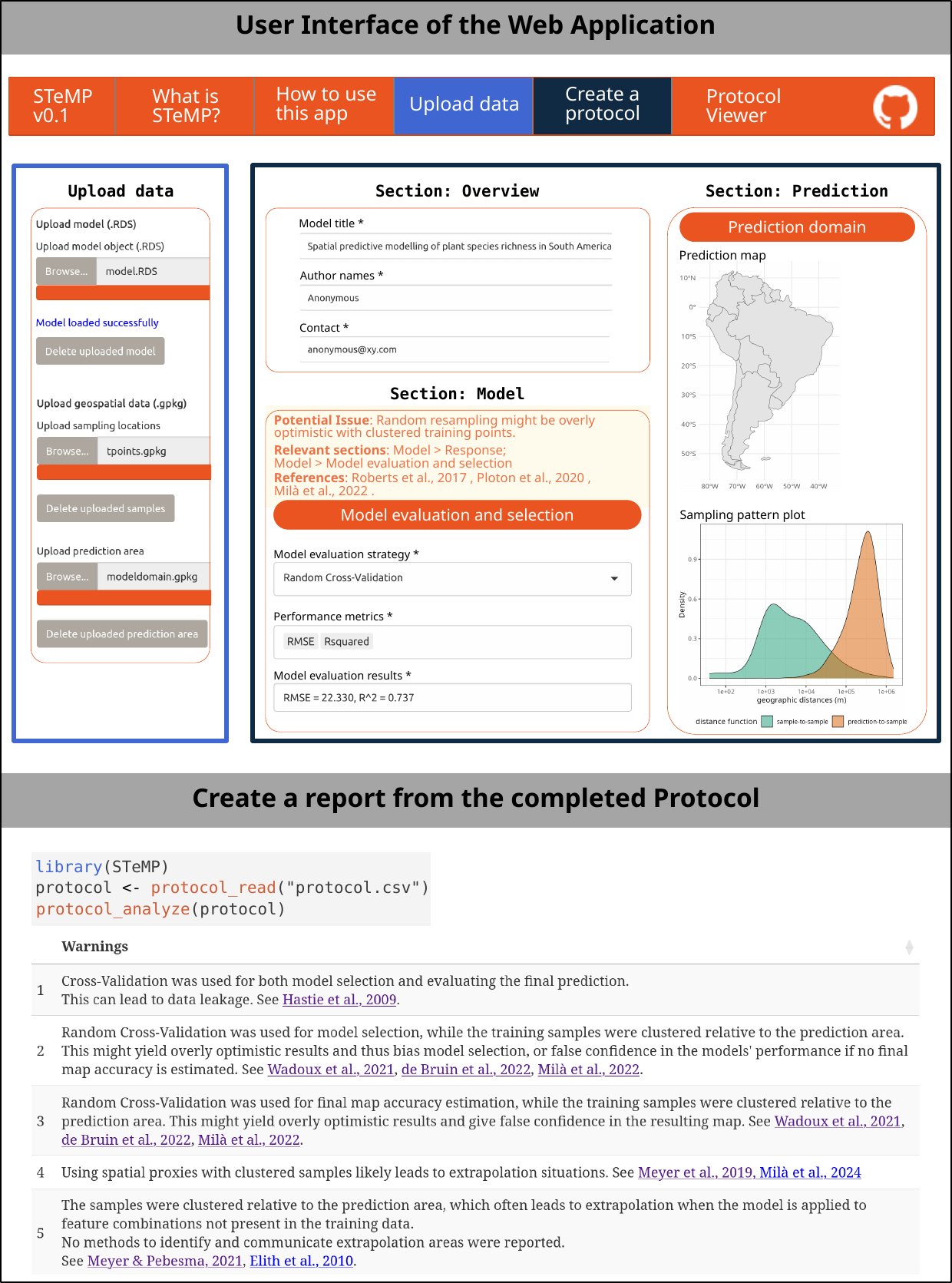}
    \caption{An overview of selected parts of the web application's user interface filled based on the example (top). The light blue rectangle shows the \emph{Upload data} tab, where the trained model, the location of the training points and the prediction area were uploaded. The dark blue rectangle shows the \emph{Create a protocol} tab, which is further divided into the panels \emph{Overview}, \emph{Model} and \emph{Prediction}. Based on the uploaded data, the \emph{Prediction} panel shows an automatically generated map of the prediction domain, as well as a plot of geographical nearest-neighbour distances. From the latter, a clustered sampling design is automatically inferred, leading to a warning message if a random cross-validation strategy is selected in the \emph{Model} panel.
    The lower part of the figure shows the \texttt{protocol\_analyze} function, which creates a report from the completed protocol.}
    \label{fig:example_app}
\end{figure}

The aim of STeMP is to facilitate a transparent and concise reporting that ultimately enables answering these open questions.
We filled the protocol using the web application. Figure \ref{fig:example_app} shows an overview over selected parts of its user interface filled with information from this example.
We uploaded the trained model, as well as geospatial data of the training locations and the prediction area in the \texttt{Upload data} tab (see the blue rectangle in Figure \ref{fig:example_app}). From these uploaded data, several fields of the protocol were automatically filled -- e.g., the evaluation results and the final hyperparameter values (see the dark blue rectangle in Figure \ref{fig:example_app}). Furthermore, maps of the training locations, the prediction area, as well as a plot comparing the nearest-neighbour distances encountered during prediction and between training points were automatically generated from the uploaded data. From comparing the nearest-neighbour distances, the web application automatically inferred the sampling design to be clustered. Since we used a random cross-validation in this example, a warning was then raised suggesting to assess the suitability of the chosen evaluation strategy for the clustered sampling design \citep[see e.g.][]{Roberts2017}.
After completing the protocol using the web application, the final protocol was downloaded as a PDF and CSV file (see Appendix). From this CSV file, we extracted all warnings that were raised by the web application using the function \texttt{protocol\_analyze} (lower part of Figure \ref{fig:example_app}).

The output of the protocol and the summary of the warnings encountered while filling the protocol now enable us to answer the questions raised before:
\begin{enumerate}
    \item The training points are clustered with regard to the prediction area (see the completed protocol in the Appendix for a map of the training points).
    \item The high $R^2$ value is likely overoptimistic when the model is applied to South America since the random cross-validation only tested the performance for prediction locations close to the training points.
    \item The spatial patterns are likely caused by using spatial proxies as predictors,  which often lead to overfitting and linear artifacts in the prediction when used with clustered training data \citep{Meyer2019}. Since no proper predictor selection strategy was employed, the protocol shows that these issues were not mitigated \citep{Meyer2019, Mila2024}. 
    \item The model was not tested for its ability to extrapolate in space, hence further testing would be needed before applying it to another region.
    \item Since the map extends over the whole of South America and the predictors have a resolution of 0.0833° ($\approx 9.3 \text{ km}$ at the equator), the map would rather be useful for continent-scale applications than for small-scale planning.
\end{enumerate}

\section{Conclusion}

STeMP enables transparent reporting of important modelling decisions and data characteristics. It is tailored towards spatio-temporal machine-learning modelling, covering characteristics such as evaluation strategies, predictor selection and uncertainty assessment in light of specific characteristics such as autocorrelation inherent in spatio-temporal data. 
Furthermore, STeMP is implemented as a web application, which not only facilitates filling the protocol, but also renders it useful even during model development by raising warnings when common pitfalls in spatio-temporal machine-learning are encountered.
Thus, STeMP fills the gap between generic model protocols and domain-specific model reports. Thereby, it covers the most important criteria as recently proposed in the vision by \citet{Graser2026}.

While STeMP covers several fields and options covering temporal aspects, it is still mainly focused on spatial machine-learning. Future developments will focus on exploring ways to further extend the protocol in the temporal dimensions. 
Furthermore, the automatic extraction of information currently only works for model objects created in R. Future work should focus on expanding this, e.g., by supporting the more generic ONNX model format \citep{ONNX}.
Additionally, the potential of large-language models is not exploited by the current protocol. This may include extracting relevant information from existing documents (e.g., research papers), and then passing this information to STeMP.
Moreover, future work may explore the possibility to automatically upload a completed protocol to Zenodo, enabling permanent storage and association with a DOI.
Lastly, the web application currently enables downloading the completed protocol as CSV and PDF file or as a ZIP folder (the latter containing the figures generated during filling the protocol). However, future work may explore exporting a JSON file, which can store information more flexibly than a CSV. 

We believe that the proposed spatio-temporal modelling protocol can act as a basis for a community-driven reporting standard, enabling transparency and fostering discussion about critical decisions and data characteristics in spatio-temporal modelling.

\section*{Acknowledgements}

The web application builds on the ODMAP protocol \citep{Zurell2020}, which standardizes reporting for species distribution models. We acknowledge the authors' foundational contribution.

This project has received the financial support of the European Union’s Horizon Europe research and innovation programme under the Marie Skłodowska-Curie grant agreement No. 101147446, as well as the project BEyond within the priority Program 1374 "Biodiversity-Exploratories" (512284513) and the TRR 391 Spatio-temporal Statistics for the Transition of Energy and Transport (520388526), both funded by the Deutsche Forschungsgemeinschaft (DFG, German Research Foundation).

\section*{Conflict of Interest statement}

The authors declare no competing interests.

\section*{Author Contributions}
JL, HM, ML and JN conceived the ideas; JL, FS, HM, ML and JN were responsible for the conceptual design of the protocol, JL and AFJ were responsible for software implementation; JL, HM and JN led the writing of the manuscript, with TK and all other authors contributing critically to the drafts. All authors gave final approval for publication.

\newpage

\appendix

\setcounter{table}{0} 
\renewcommand{\thetable}{A\arabic{table}} 

\input{table}

\bibliographystyle{unsrtnat}
\bibliography{literature} 

\end{document}

%% file: table.tex
{
\footnotesize
\renewcommand{\arraystretch}{1.1} 
\begin{xltabular}{\textwidth}{l l X}
\caption{The sections and fields of STeMP. Mandatory fields are shown in italic.} \label{tab:model_elements} \\
\toprule
\textbf{Section} & \textbf{Field} & \textbf{Details} \\
\midrule
\endfirsthead

\toprule
\textbf{Section} & \textbf{Field} & \textbf{Details} \\
\midrule
\endhead

\bottomrule
\endfoot

\bottomrule
\endlastfoot

\multirow{7}{*}{\textbf{Overview}} 
& \textit{Model title} & The title of the model \\
& \textit{Author names} & The names of the authors \\
& \textit{Contact} & Contact email address \\
& Study title & The title of the study associated with the model \\
& Study link & The link to the study associated with the model \\
& \textit{Target variable} & The response variable the model aims to predict \\
& \textit{Scientific Field} & The scientific field in which the model was developed \\
\midrule

\multirow{36}{*}{\textbf{Model}} 
& \textit{Model type} & Type of the model (e.g., classification, regression) \\
& \textit{Algorithm} & Name of the applied algorithm \\
& Architecture & Description of the model's architecture (nodes, layers etc.) \\
\cmidrule{2-3}
& \textit{Sampling locations} & Map of training locations \\
& \textit{Sampling extent} & The spatial/temporal extent at which the training points were collected \\
& \textit{Sample acquisition} & Method to collect the training points (e.g., field sampling) \\
& Sample geometry & The geometry type of the training points (e.g., Polygons, Points) \\
& \textit{Sample size} & The number of training points \\
& Classes & The number of response classes \\
& Samples per class & The number of training points in each class \\
& Range & The range of response values represented in the training data \\
& \textit{Sampling pattern} & The sampling pattern ranging from random to clustered \\
& Coordinate Reference System & The coordinate reference system used \\
& Data sources of response & The source(s) of the response data \\
\cmidrule{2-3}
& \textit{Predictor types} & Types of predictor data used to model the response \\
& \textit{Number of predictors} & The initial number of predictor variables \\
& Names of predictors & List the names of all initial predictor variables \\
& Resolution of predictors & The spatial resolution and temporal frequency/grain of the predictors \\
& Preprocessing & Detail on preprocessing steps \\
& Temporal alignment & Temporal alignment of the predictors with the response \\
& Data sources of predictors & Data source(s) of the predictors \\
\cmidrule{2-3}
& \textit{Model evaluation strategy} & Details on the evaluation strategy used for model selection (e.g., random cross-validation, spatial block cross-validation, temporal block cross-validation) \\
& \textit{Performance metrics} & Performance metric(s) used during model evaluation \\
& \textit{Model evaluation results} & Results of the model evaluation \\
& \textit{Hyperparameter tuning} & Hyperparameter tuning methods (e.g., grid search) \\
& \textit{Predictor selection} & Predictor selection methods (e.g., forward-feature selection) \\
\cmidrule{2-3}
& Explainability & Methods of explainable AI used to gain insights into the models behavior \\
& Scientific interpretation & Discussion of the implications of the model in the context of theory \\
\cmidrule{2-3}
& \textit{Potential biases} & Specifications of potential biases in the data or algorithm \\
& Sensitivity assessment & Assessment of the impact of varying input parameters \\
& \textit{Limitations} & Descriptions of limitations when applying the model \\
\cmidrule{2-3}
& \textit{Software} & Programming language and versioning of key software libraries \\
& \textit{Code availability} & Code availability (e.g., link to code repository) \\
& \textit{Data availability} & Data availability (e.g., link to data repository) \\
\midrule

\multirow{10}{*}{\textbf{Prediction}}
& \textit{Prediction extent} & Spatial/Temporal extent of the prediction domain \\
& \textit{Prediction resolution} & The spatial resolution and temporal frequency/grain of the prediction domain \\
\cmidrule{2-3}
& \textit{Map evaluation strategy} & Details on the evaluation of the predicted map \\
& \textit{Performance metrics} & Performance metrics used to evaluate the predicted map \\
& \textit{Map accuracy} & Estimated map accuracy \\
& Uncertainty quantification & Communication of prediction uncertainty (e.g., area of applicability) \\
\cmidrule{2-3}
& Threshold selection & Details on threshold selection if the prediction was binarized \\
& Post-processing & Post-processing of the prediction (e.g., spatial filters to smooth the prediction) \\
\end{xltabular}
}

%% file: literature.bib
@article{Zurell2020,
	title = {A standard protocol for reporting species distribution models},
	volume = {43},
	issn = {0906-7590, 1600-0587},
	url = {https://onlinelibrary.wiley.com/doi/10.1111/ecog.04960},
	doi = {10.1111/ecog.04960},
	language = {en},
	number = {9},
	urldate = {2024-01-17},
	journal = {Ecography},
	author = {Zurell, Damaris and Franklin, Janet and König, Christian and Bouchet, Phil J. and Dormann, Carsten F. and Elith, Jane and Fandos, Guillermo and Feng, Xiao and Guillera‐Arroita, Gurutzeta and Guisan, Antoine and Lahoz‐Monfort, José J. and Leitão, Pedro J. and Park, Daniel S. and Peterson, A. Townsend and Rapacciuolo, Giovanni and Schmatz, Dirk R. and Schröder, Boris and Serra‐Diaz, Josep M. and Thuiller, Wilfried and Yates, Katherine L. and Zimmermann, Niklaus E. and Merow, Cory},
	month = sep,
	year = {2020},
	pages = {1261--1277},
}

@article{Grimm2006,
	title = {A standard protocol for describing individual-based and agent-based models},
	volume = {198},
	issn = {03043800},
	url = {https://linkinghub.elsevier.com/retrieve/pii/S0304380006002043},
	doi = {10.1016/j.ecolmodel.2006.04.023},
	language = {en},
	number = {1-2},
	urldate = {2024-01-17},
	journal = {Ecological Modelling},
	author = {Grimm, Volker and Berger, Uta and Bastiansen, Finn and Eliassen, Sigrunn and Ginot, Vincent and Giske, Jarl and Goss-Custard, John and Grand, Tamara and Heinz, Simone K. and Huse, Geir and Huth, Andreas and Jepsen, Jane U. and Jørgensen, Christian and Mooij, Wolf M. and Müller, Birgit and Pe’er, Guy and Piou, Cyril and Railsback, Steven F. and Robbins, Andrew M. and Robbins, Martha M. and Rossmanith, Eva and Rüger, Nadja and Strand, Espen and Souissi, Sami and Stillman, Richard A. and Vabø, Rune and Visser, Ute and DeAngelis, Donald L.},
	month = sep,
	year = {2006},
	pages = {115--126},
}

@inproceedings{Mitchell2019,
	address = {Atlanta GA USA},
	title = {Model {Cards} for {Model} {Reporting}},
	isbn = {978-1-4503-6125-5},
	url = {https://dl.acm.org/doi/10.1145/3287560.3287596},
	doi = {10.1145/3287560.3287596},
	language = {en},
	urldate = {2024-01-17},
	booktitle = {Proceedings of the {Conference} on {Fairness}, {Accountability}, and {Transparency}},
	publisher = {ACM},
	author = {Mitchell, Margaret and Wu, Simone and Zaldivar, Andrew and Barnes, Parker and Vasserman, Lucy and Hutchinson, Ben and Spitzer, Elena and Raji, Inioluwa Deborah and Gebru, Timnit},
	month = jan,
	year = {2019},
	pages = {220--229},
}

@article{Seuru2026,
	title = {The {ODE} ({Overview}, {Data}, and {Execution}) protocol for a standardized use of machine learning in environmental, social and related interdisciplinary sciences},
	volume = {198},
	issn = {13648152},
	url = {https://linkinghub.elsevier.com/retrieve/pii/S1364815226000599},
	doi = {10.1016/j.envsoft.2026.106912},
	urldate = {2026-02-20},
	journal = {Environmental Modelling \& Software},
	author = {Seuru, Samuel and Grimm, Volker and Barton, Michael and Perez, Liliana and Mahdizadeh Gharakhanlou, Navid and Sengupta, Raja and Dagnino, Alejandro Miguel},
	month = mar,
	year = {2026},
	pages = {106912},
}

@Manual{Fay2026,
  title = {golem: A Framework for Robust Shiny Applications},
  author = {Colin Fay and Vincent Guyader and Sébastien Rochette and Cervan Girard},
  year = {2026},
  note = {R package version 0.5.1.9015},
  url = {https://thinkr-open.github.io/golem/},
}

@book{Fay2021,
author = {Fay, Colin and Rochette, Sébastien and Guyader, Vincent and Girard, Cervan},
year = {2021},
month = {09},
pages = {},
title = {Engineering Production-Grade Shiny Apps},
isbn = {9781003029878},
doi = {10.1201/9781003029878}
}

@book{Wickham2021,
  title={Mastering Shiny: Build Interactive Apps, Reports, and Dashboards Powered by R},
  author={Wickham, Hadley},
  isbn={9781492047384},
  lccn={2022301399},
  url={https://books.google.de/books?id=nrvAzQEACAAJ},
  year={2021},
  publisher={O'Reilly}
}

@Manual{Chang2026,
  title = {shiny: Web Application Framework for R},
  author = {Winston Chang and Joe Cheng and JJ Allaire and Carson Sievert and Barret Schloerke and Garrick Aden-Buie and Yihui Xie and Jeff Allen and Jonathan McPherson and Alan Dipert and Barbara Borges},
  year = {2026},
  note = {R package version 1.13.0.9000},
  url = {https://shiny.posit.co/},
}

@Article{Wickham2011,
  author = {Hadley Wickham},
  title = {testthat: Get Started with Testing},
  journal = {The R Journal},
  year = {2011},
  volume = {3},
  pages = {5--10},
  url = {https://journal.r-project.org/articles/RJ-2011-002/},
}

@Manual{Chang2024,
  title = {shinytest: Test Shiny Apps},
  author = {Winston Chang and Gábor Csárdi and Hadley Wickham},
  year = {2024},
  note = {R package version 1.6.0},
  url = {https://github.com/rstudio/shinytest},
}

@article{Wadoux2025,
author = {Wadoux, Alexandre M. J.-C.},
title = {Artificial intelligence in soil science},
journal = {European Journal of Soil Science},
volume = {76},
number = {2},
pages = {e70080},
doi = {https://doi.org/10.1111/ejss.70080},
url = {https://bsssjournals.onlinelibrary.wiley.com/doi/abs/10.1111/ejss.70080},
eprint = {https://bsssjournals.onlinelibrary.wiley.com/doi/pdf/10.1111/ejss.70080},
note = {e70080 EJSS-594-24.R2},
year = {2025}
}

@article{Pichler2023,
author = {Pichler, Maximilian and Hartig, Florian},
title = {Machine learning and deep learning—A review for ecologists},
journal = {Methods in Ecology and Evolution},
volume = {14},
number = {4},
pages = {994-1016},
doi = {https://doi.org/10.1111/2041-210X.14061},
url = {https://besjournals.onlinelibrary.wiley.com/doi/abs/10.1111/2041-210X.14061},
eprint = {https://besjournals.onlinelibrary.wiley.com/doi/pdf/10.1111/2041-210X.14061},
year = {2023}
}

@article{Yang2024,
title = {Interpretable machine learning for weather and climate prediction: A review},
journal = {Atmospheric Environment},
volume = {338},
pages = {120797},
year = {2024},
issn = {1352-2310},
doi = {https://doi.org/10.1016/j.atmosenv.2024.120797},
url = {https://www.sciencedirect.com/science/article/pii/S1352231024004722},
author = {Ruyi Yang and Jingyu Hu and Zihao Li and Jianli Mu and Tingzhao Yu and Jiangjiang Xia and Xuhong Li and Aritra Dasgupta and Haoyi Xiong},
}

@article{Mongan2020,
	title = {Checklist for {Artificial} {Intelligence} in {Medical} {Imaging} ({CLAIM}): {A} {Guide} for {Authors} and {Reviewers}},
	volume = {2},
	issn = {2638-6100},
	shorttitle = {Checklist for {Artificial} {Intelligence} in {Medical} {Imaging} ({CLAIM})},
	url = {http://pubs.rsna.org/doi/10.1148/ryai.2020200029},
	doi = {10.1148/ryai.2020200029},
	language = {en},
	number = {2},
	urldate = {2025-05-26},
	journal = {Radiology: Artificial Intelligence},
	author = {Mongan, John and Moy, Linda and Kahn, Charles E.},
	month = mar,
	year = {2020},
	pages = {e200029},
}

@article{Luo2016,
	title = {Guidelines for {Developing} and {Reporting} {Machine} {Learning} {Predictive} {Models} in {Biomedical} {Research}: {A} {Multidisciplinary} {View}},
	volume = {18},
	issn = {1438-8871},
	shorttitle = {Guidelines for {Developing} and {Reporting} {Machine} {Learning} {Predictive} {Models} in {Biomedical} {Research}},
	url = {http://www.jmir.org/2016/12/e323/},
	doi = {10.2196/jmir.5870},
	language = {en},
	number = {12},
	urldate = {2025-05-26},
	journal = {Journal of Medical Internet Research},
	author = {Luo, Wei and Phung, Dinh and Tran, Truyen and Gupta, Sunil and Rana, Santu and Karmakar, Chandan and Shilton, Alistair and Yearwood, John and Dimitrova, Nevenka and Ho, Tu Bao and Venkatesh, Svetha and Berk, Michael},
	month = dec,
	year = {2016},
	pages = {e323},
}

@article{Collins2015,
	title = {Transparent reporting of a multivariable prediction model for individual prognosis or diagnosis ({TRIPOD}): the {TRIPOD} {Statement}},
	volume = {13},
	issn = {1741-7015},
	shorttitle = {Transparent reporting of a multivariable prediction model for individual prognosis or diagnosis ({TRIPOD})},
	url = {http://www.biomedcentral.com/1741-7015/13/1},
	doi = {10.1186/s12916-014-0241-z},
	language = {en},
	number = {1},
	urldate = {2025-05-26},
	journal = {BMC Medicine},
	author = {Collins, Gary S and Reitsma, Johannes B and Altman, Douglas G and Moons, Karel},
	year = {2015},
	pages = {1},
}

@ARTICLE{Ram2019,
  author={Ram, Karthik and Boettiger, Carl and Chamberlain, Scott and Ross, Noam and Salmon, Maëlle and Butland, Stefanie},
  journal={Computing in Science \& Engineering},
  title={A Community of Practice Around Peer Review for Long-Term Research Software Sustainability}, 
  year={2019},
  volume={21},
  number={2},
  pages={59-65},
  doi={10.1109/MCSE.2018.2882753}}

@Article{Linnenbrink2024,
AUTHOR = {Linnenbrink, Jan and Mil\`a, Carles and Ludwig, Marvin and Meyer, Hanna},
TITLE = {kNNDM CV: $k$-fold nearest-neighbour distance matching cross-validation for map accuracy estimation},
JOURNAL = {Geoscientific Model Development},
VOLUME = {17},
YEAR = {2024},
NUMBER = {15},
PAGES = {5897--5912},
URL = {https://gmd.copernicus.org/articles/17/5897/2024/},
DOI = {10.5194/gmd-17-5897-2024}
}

@Article{Roberts2017,
  author    = {Roberts, David R. and Bahn, Volker and Ciuti, Simone and Boyce, Mark S. and Elith, Jane and Guillera-Arroita, Gurutzeta and Hauenstein, Severin and Lahoz-Monfort, Jos\'es J. and Schr{\"o}der, Boris and Thuiller, Wilfried and Warton, David I. and Wintle, Brendan A. and Hartig, Florian and Dormann, Carsten F.},
  title     = {{C}ross-validation strategies for data with temporal, spatial, hierarchical, or phylogenetic structure},
  journal   = {Ecography},
  year      = {2017},
  issn      = {1600-0587},
  doi       = {10.1111/ecog.02881},
  publisher = {Blackwell Publishing Ltd},
}

@Inbook{Meyer2026,
author="Meyer, Hanna
and Ludwig, Marvin
and Mil{\`a}, Carles
and Linnenbrink, Jan
and Schumacher, Fabian",
editor="Rocchini, Duccio",
title="The CAST Package for Training and Assessment of Spatial Prediction Models",
bookTitle="R Coding for Ecology",
year="2026",
publisher="Springer Nature Switzerland",
address="Cham",
pages="247--266",
isbn="978-3-031-99665-8",
doi="10.1007/978-3-031-99665-8_11",
url="https://doi.org/10.1007/978-3-031-99665-8_11"
}

@book{Baddeley2015,
  title={Spatial point patterns: methodology and applications with R},
  author={Baddeley, Adrian and Rubak, Ege and Turner, Rolf},
  year={2015},
  publisher={CRC press}
}

@article{Sabatini2021,
author = {Sabatini, Francesco Maria and Lenoir, Jonathan and Hattab, Tarek and Arnst, Elise Aimee and Chytrý, Milan and Dengler, Jürgen and De Ruffray, Patrice and Hennekens, Stephan M. and Jandt, Ute and Jansen, Florian and Jiménez-Alfaro, Borja and Kattge, Jens and Levesley, Aurora and Pillar, Valério D. and Purschke, Oliver and Sandel, Brody and Sultana, Fahmida and Aavik, Tsipe and Aćić, Svetlana and Acosta, Alicia T. R. and Agrillo, Emiliano and Alvarez, Miguel and Apostolova, Iva and Arfin Khan, Mohammed A. S. and Arroyo, Luzmila and Attorre, Fabio and Aubin, Isabelle and Banerjee, Arindam and Bauters, Marijn and Bergeron, Yves and Bergmeier, Erwin and Biurrun, Idoia and Bjorkman, Anne D. and Bonari, Gianmaria and Bondareva, Viktoria and Brunet, Jörg and Čarni, Andraž and Casella, Laura and Cayuela, Luis and Černý, Tomáš and Chepinoga, Victor and Csiky, János and Ćušterevska, Renata and De Bie, Els and de Gasper, André Luis and De Sanctis, Michele and Dimopoulos, Panayotis and Dolezal, Jiri and Dziuba, Tetiana and El-Sheikh, Mohamed Abd El-Rouf Mousa and Enquist, Brian and Ewald, Jörg and Fazayeli, Farideh and Field, Richard and Finckh, Manfred and Gachet, Sophie and Galán-de-Mera, Antonio and Garbolino, Emmanuel and Gholizadeh, Hamid and Giorgis, Melisa and Golub, Valentin and Alsos, Inger Greve and Grytnes, John-Arvid and Guerin, Gregory Richard and Gutiérrez, Alvaro G. and Haider, Sylvia and Hatim, Mohamed Z. and Hérault, Bruno and Hinojos Mendoza, Guillermo and Hölzel, Norbert and Homeier, Jürgen and Hubau, Wannes and Indreica, Adrian and Janssen, John A. M. and Jedrzejek, Birgit and Jentsch, Anke and Jürgens, Norbert and Kącki, Zygmunt and Kapfer, Jutta and Karger, Dirk Nikolaus and Kavgacı, Ali and Kearsley, Elizabeth and Kessler, Michael and Khanina, Larisa and Killeen, Timothy and Korolyuk, Andrey and Kreft, Holger and Kühl, Hjalmar S. and Kuzemko, Anna and Landucci, Flavia and Lengyel, Attila and Lens, Frederic and Lingner, Débora Vanessa and Liu, Hongyan and Lysenko, Tatiana and Mahecha, Miguel D. and Marcenò, Corrado and Martynenko, Vasiliy and Moeslund, Jesper Erenskjold and Monteagudo Mendoza, Abel and Mucina, Ladislav and Müller, Jonas V. and Munzinger, Jérôme and Naqinezhad, Alireza and Noroozi, Jalil and Nowak, Arkadiusz and Onyshchenko, Viktor and Overbeck, Gerhard E. and Pärtel, Meelis and Pauchard, Aníbal and Peet, Robert K. and Peñuelas, Josep and Pérez-Haase, Aaron and Peterka, Tomáš and Petřík, Petr and Peyre, Gwendolyn and Phillips, Oliver L. and Prokhorov, Vadim and Rašomavičius, Valerijus and Revermann, Rasmus and Rivas-Torres, Gonzalo and Rodwell, John S. and Ruprecht, Eszter and Rūsiņa, Solvita and Samimi, Cyrus and Schmidt, Marco and Schrodt, Franziska and Shan, Hanhuai and Shirokikh, Pavel and Šibík, Jozef and Šilc, Urban and Sklenář, Petr and Škvorc, Željko and Sparrow, Ben and Sperandii, Marta Gaia and Stančić, Zvjezdana and Svenning, Jens-Christian and Tang, Zhiyao and Tang, Cindy Q. and Tsiripidis, Ioannis and Vanselow, Kim André and Vásquez Martínez, Rodolfo and Vassilev, Kiril and Vélez-Martin, Eduardo and Venanzoni, Roberto and Vibrans, Alexander Christian and Violle, Cyrille and Virtanen, Risto and von Wehrden, Henrik and Wagner, Viktoria and Walker, Donald A. and Waller, Donald M. and Wang, Hua-Feng and Wesche, Karsten and Whitfeld, Timothy J. S. and Willner, Wolfgang and Wiser, Susan K. and Wohlgemuth, Thomas and Yamalov, Sergey and Zobel, Martin and Bruelheide, Helge},
title = {sPlotOpen – An environmentally balanced, open-access, global dataset of vegetation plots},
journal = {Global Ecology and Biogeography},
volume = {30},
number = {9},
pages = {1740-1764},
doi = {https://doi.org/10.1111/geb.13346},
url = {https://onlinelibrary.wiley.com/doi/abs/10.1111/geb.13346},
eprint = {https://onlinelibrary.wiley.com/doi/pdf/10.1111/geb.13346},
year = {2021}
}

@article{Fick2017,
author = {Fick, Stephen E. and Hijmans, Robert J.},
title = {WorldClim 2: new 1-km spatial resolution climate surfaces for global land areas},
journal = {International Journal of Climatology},
volume = {37},
number = {12},
pages = {4302-4315},
doi = {https://doi.org/10.1002/joc.5086},
url = {https://rmets.onlinelibrary.wiley.com/doi/abs/10.1002/joc.5086},
eprint = {https://rmets.onlinelibrary.wiley.com/doi/pdf/10.1002/joc.5086},
year = {2017}
}

@article{Graser2026,
	title = {Spatiotemporal {Model} {Cards} {Enabling} {Future}-{Proof} {GeoAI} {Systems}},
	volume = {12},
	issn = {2374-0353, 2374-0361},
	url = {https://dl.acm.org/doi/10.1145/3809183},
	doi = {10.1145/3809183},
	language = {en},
	number = {2},
	urldate = {2026-05-19},
	journal = {ACM Transactions on Spatial Algorithms and Systems},
	author = {Graser, Anita and Wachsenegger, Anahid and Doulkeridis, Christos and Theodoropoulos, George and Salehi, Bahare and Dragaschnig, Melitta and Antoniou, Stathis and Theodoridis, Yannis},
	month = jun,
	year = {2026},
	pages = {1--39},
}

@Manual{tidymodels,
    title = {Tidymodels: a collection of packages for modeling and machine learning using tidyverse principles.},
    author = {Max Kuhn and Hadley Wickham},
    url = {https://www.tidymodels.org},
    year = {2020},
}

@Article{caret,
    title = {Building Predictive Models in R Using the caret Package},
    volume = {28},
    url = {https://www.jstatsoft.org/index.php/jss/article/view/v028i05},
    doi = {10.18637/jss.v028.i05},
    number = {5},
    journal = {Journal of Statistical Software},
    author = {Max Kuhn},
    year = {2008},
    pages = {1–26},
}

@Article{mlr3,
    title = {{mlr3}: A modern object-oriented machine learning framework in {R}},
    author = {Michel Lang and Martin Binder and Jakob Richter and Patrick Schratz and Florian Pfisterer and Stefan Coors and Quay Au and Giuseppe Casalicchio and Lars Kotthoff and Bernd Bischl},
    journal = {Journal of Open Source Software},
    year = {2019},
    month = {dec},
    doi = {10.21105/joss.01903},
    url = {https://joss.theoj.org/papers/10.21105/joss.01903},
}

@article{Meyer2019,
title = {Importance of spatial predictor variable selection in machine learning applications – Moving from data reproduction to spatial prediction},
journal = {Ecological Modelling},
volume = {411},
pages = {108815},
year = {2019},
issn = {0304-3800},
doi = {https://doi.org/10.1016/j.ecolmodel.2019.108815},
url = {https://www.sciencedirect.com/science/article/pii/S0304380019303230},
author = {Hanna Meyer and Christoph Reudenbach and Stephan Wöllauer and Thomas Nauss},
}

@Article{Mila2024,
AUTHOR = {Mil\`a, Carles and Ludwig, Marvin and Pebesma, Edzer and Tonne, Cathryn and Meyer, Hanna},
TITLE = {Random forests with spatial proxies for environmental modelling: opportunities and pitfalls},
JOURNAL = {EGUsphere},
VOLUME = {2024},
YEAR = {2024},
PAGES = {1--30},
URL = {https://egusphere.copernicus.org/preprints/2024/egusphere-2024-138/},
DOI = {10.5194/egusphere-2024-138}
}

@article{Meyer2021,
author = {Meyer, Hanna and Pebesma, Edzer},
title = {Predicting into unknown space? Estimating the area of applicability of spatial prediction models},
journal = {Methods in Ecology and Evolution},
volume = {12},
number = {9},
pages = {1620-1633},
doi = {https://doi.org/10.1111/2041-210X.13650},
url = {https://besjournals.onlinelibrary.wiley.com/doi/abs/10.1111/2041-210X.13650},
year = {2021}
}

@article{Elith2010,
author = {Elith, Jane and Kearney, Michael and Phillips, Steven},
title = {The art of modelling range-shifting species},
journal = {Methods in Ecology and Evolution},
volume = {1},
number = {4},
pages = {330-342},
doi = {https://doi.org/10.1111/j.2041-210X.2010.00036.x},
url = {https://besjournals.onlinelibrary.wiley.com/doi/abs/10.1111/j.2041-210X.2010.00036.x},
year = {2010}
}

@misc{ONNX,
    author = {Bai, Junjie and Lu, Fang and Zhang, Ke and others},
    title = {ONNX: Open Neural Network Exchange},
    year = {2025},
    publisher = {GitHub},
    journal = {GitHub repository},
    howpublished = {\url{https://github.com/onnx/onnx}}
}

@article{Meinshausen2006,
  title={Quantile regression forests.},
  author={Meinshausen, Nicolai and Ridgeway, Greg},
  journal={Journal of machine learning research},
  volume={7},
  number={6},
  year={2006}
}

@article{Nowosad2026,
    title={Navigating challenges in spatial machine learning: Validation, uncertainty, algorithms, and reproducibility}, 
    url={https://www.erdkunde.uni-bonn.de/article/view/3471}, 
    DOI={10.3112/erdkunde.2026.04.01}, 
    journal={ERDKUNDE}, 
    author={Nowosad, Jakub and Bonannella, Carmelo and Görgen, Darius and Jemeljanova, Marta and Kattenborn, Teja and Linnenbrink, Jan and Meyer, Hanna and Nussbaum, Madlene and Patelli, Luca and Simoes, Rolf and Uuemaa, Evelyn}, 
    year={2026}, 
    month={Jun.} }

@article{Jarvis2008,
author = {Jarvis, Andrew and Reuter, Hannes and Nelson, Andy and Guevara, Edith},
year = {2008},
month = {01},
pages = {},
title = {Hole-filled seamless SRTM data v4},
journal = {International Centre for Tropical Agriculture (CIAT)}
}
